# ENHANCEMENT TECHNIQUES FOR LOCAL CONTENT PRESERVATION AND CONTRAST IMPROVEMENT IN IMAGES

Chelsy Sapna Josephus and Remya .S

Department of Computer Science, University of Kerala, Thiruvananthapuram, India
chelze4u@gmail.com , remyayes@gmail.com

**ABSTRACT**

There are several images that do not have uniform brightness which pose a challenging problem for image enhancement systems. As histogram equalization has been successfully used to correct for uniform brightness problems, a histogram equalization method that utilizes human visual system based thresholding(human vision thresholding) as well as logarithmic processing techniques were introduced later . But these methods are not good for preserving the local content of the image which is a major factor for various images like medical and aerial images. Therefore new method is proposed here. This method is referred as "Human vision thresholding with enhancement technique for dark blurred images for local content preservation". It uses human vision thresholding together with an existing enhancement method for dark blurred images. Furthermore a comparative study with another method for local content preservation is done which is further extended to make it suitable for contrast improvement . Experimental results shows that the proposed methods outperforms the former existing methods in preserving the local content for standard images ,medical and aerial images .

## 1. INTRODUCTION

Digital image enhancement is necessary to improve the visual appearance of the image .It is also used to provide a better transform representation for future automated image processing such as image analysis, segmentation and recognition [1-2]. To discern the concealed but important information in the images, it is needed that the usage of various image enhancement methods such as enhancing edges, emphasizing the differences, or reducing the noise is done. Processing techniques for image enhancement can be classified into spatial domain enhancement and transform domain enhancement methods. Adaptive histogram equalization(AHE) [3] belongs to spatially non uniform enhancement technique. Spatial domain enhancement techniques deal with the image's direct intensity values. While the spatially uniform methods use a transformation applied to all pixels of the image, the later methods use an input output transformation that varies adaptively with the local characteristics of the image. Histogram equalization suffers from the problem of being poorly suited for retaining local detail due to its global treatment of the image. Histogram equalization tends to over-enhance the image contrast if there is a high peaks in the histogram resulting in a undesired loss of visual data, of quality and of intensity scale . Also small scale details that are often associated with the small bins of the histogram are eliminated. AHE applies locally varying gray-scale transformation on each small region of the image. This method does not completely eliminate noise enhancement in smooth regions.

Local histogram equalization (LHE) is one of the popular branches for local image enhancement. Generally, LHE uses a small window to define a contextual region (CR) for the centre pixel of that window .Only the block of pixels that fall in this window is taken into the account for the calculation of cumulative density function(CDF). Therefore, as the window slides, the CDF is modified. The CDF is the main contributor for the LHE transform function. Hence, in LHE, the transform function of a pixel is depending on the statistics of its neighbors in CR. Because the transform function changes as a response to the changes in the contents of CR, LHE is also popularly known as adaptive histogram equalization (AHE).

Multi histogram Equalization is another efficient way of enhancing an image. But Multi-histogram equalization has generally been limited to bi-histogram equalization, and it has been previously proved that tri-histogram equalization does not have any consistent advantage. So in the previous methods the introduction of the human visual system based multi-histogram equalization method is done . This algorithm focusses on the advantages of multi-histogram equalization with the benefit of an effective quantitative measure to ensure optimal results while removing useless information to avoid the production of unwanted artifacts.This method overcomes this limitation by separating the image into different regions of illumination instead of thresholding by simple pixel intensity values. In this manner, histogram equalization can be done on each region to correct for non uniform illumination. In order to perform this segmentation, model of the human visual system is utilized . HVS-based image enhancement aims to specify the way in which the HVS discriminates between useful and unwanted data.

.

## 2. METHODS

### 2.1 HUMAN VISUAL SYSTEM BASED IMAGE ENHANCEMENT(HVS)[12]

Human Visual System (HVS) based image Enhancement aims to emulate the way in which the human visual system diffferentiates between useful and unwanted data and it is based on the background illumination and the gradient of an image . The former is arrived at using the following formula:

$$B(x,y) = \left[\frac{1}{2}\left(\frac{1}{4}\sum_Q X(i,j) + \frac{1}{4\sqrt{2}}\sum_{Q'} X(k,l)\right) + X(x,y)\right] + 2 \quad \ldots(1)$$

where B(x,y) is the background intensity at each pixel, X(x,y) is the input image, Q is all of the pixels which are directly up, down, left, and right from the pixel, and Q' is all of the pixels diagonally one pixel away. Here a parameter BT is also defined, which is the difference in the maximum and minimum graylevels of an image , arrived at using:

$BT = max(X(x, y)) - min(X(x, y))\ldots\ldots\ldots\ldots\ldots\ldots\ldots\ldots\ldots\ldots\ldots\ldots(2)$

Further, the gradient information is also needed , which is arrived at in the following formula:
$G1 = X(x,y) - X(x,y+1)\ldots\ldots\ldots\ldots\ldots\ldots\ldots\ldots\ldots\ldots\ldots\ldots\ldots(3)$
$G2 = X(x,y) \; X(x+1,y)\ldots\ldots\ldots\ldots\ldots\ldots\ldots\ldots\ldots\ldots\ldots\ldots\ldots(4)$

$$X'(x,y) = (|G1| + |G2|)/2 \quad \quad \quad (5)$$

where X'(x,y) is the gradient information and G1, G2 are the directional gradients. Finally, we must also take into consideration some parameters concerning the human eye itself, which is referred as Bxi, i= 1,2,3 and Ki, i= 1,2,3. These are arrived at using the following formulas:

$$B_{x1} = \alpha_1 B_T \quad \quad \quad (6)$$

$$B_{x2} = \alpha_2 B_T \quad \quad \quad (7)$$

$$B_{x3} = \alpha_3 B_T \quad \quad \quad (8)$$

$$K1 = \frac{1}{100} * \beta * \max\left(\frac{X'(x,y)}{B(x,y)}\right) \quad \quad \quad (9)$$

$$K2 = K1\sqrt{B_{x2}} \quad \quad \quad (10)$$

$$K3 = \frac{K1}{B_{x3}} \quad \quad \quad (11)$$

Where $\alpha_1$, $\alpha_2$, $\alpha_3$ are parameters based upon the three distinct regions of response characteristics displayed by the human eye. As a, is the lower saturation level, it is effective to set this to 0. Here $\alpha_2$, $\alpha_3$, is determined experimentally.

Using this information, the image is first broken up into the different regions of human visual response. These different regions are characterized by the minimum difference between two pixel intensities for the human visual system to register a difference. The next step is to threshold the three regions, removing the pixels which do not constitute a noticeable change for a human observer and placing these in a fourth image. These four images are arrived at using the following formula:

Im1=X(x,y) such that $Bx2 \geq B(x,y) \geq Bx1$ & $\dfrac{X'(x,y)}{\sqrt{B(x,y)}} \geq K2$ .......(12)

Im2=X(x,y) such that $Bx3 \geq B(x,y) \geq Bx2$ & $\dfrac{X'(x,y)}{B(x,y)} \geq K1$ .......(13)

Im3=X(x,y) such that $B(x,y) \geq Bx3$ & $\dfrac{X'(x,y)}{(B(x,y))^2} \geq K3$ ...........(14)

Im4=X(x,y)   All Remaining pixels

These four images are then enhanced separately and recombined to form the enhanced image. The above mentioned thresholding method is also referred as human vision thresholding.

HVS based histogram Equalization is a method that uses the concept of multi-histogram equalization. By separating the image into regions by the quality of illumination, such as over-illuminated, well illuminated, and under-illuminated, traditional histogram equalization can be used on each region to correct for non-uniform illumination. For this, the utilization of the human visual system to segment the image, using the measure of image enhancement to select $\alpha_2$ and $\alpha_3$ is done. The first three images are then equalized separately and unionized, with the remaining pixels filled in. In summary, the algorithm is executed as follows:

1. Obtain the input image and segment image using Human Vision Thresholding algorithm
2. Equalize the three images separately.
3. Recombine the pixels in the three equalized images
4. Fill in the missing pixels to generate the output image.

## 2.2 HUMAN VISION BASED SEGMENTATION WITH EDGE PRESERVING CONTRAST ENHANCEMENT(EPCE) (HVSedge)[14]

This is another method which uses the human vision based thresholding for segmentation. This method is a variation of HVS method. Here unlike the HVS method, the images Im1, Im2 and Im4 satisfying the corresponding thresholding conditions as in HVS method is considered. Then one of the images obtained is enhanced by edge preserving enhancement method and the remaining images are enhanced using classical histogram equalization method.

EPCE(Edge preserving contrast enhancement) is a enhancement algorithm which is designed to preserve edges while improving contrast locally by combining the output of an edge-detection algorithm with the original spatial information of the image. This achieves a more robust enhancement algorithm that is able to perform edge detection or enhancement. This enhancement algorithm can be used with any suitable edge-detection algorithm. It uses preprocessing steps to standardize image brightness and several post processing steps to enhance the edges contained.

**Step1 :**
Initially the following operation is done on each image pixel, using the following formula (eqn 15). This operation is based on the local mean at each image pixel.

$$I(x, y) = \frac{2}{1 + e^{2\tau(x,y)/\lambda(x,y)}} - 1 \quad \ldots\ldots\ldots\ldots\ldots\ldots(15)$$

where I(x, y) is the output image, τ (x, y) is the gray scale image, and λ is the local statistic of the image used to adjust the transfer function to the local mean. Finally, λ is

$$\lambda(x, y) = C + (M - C)\left(\frac{\mu(x, y)}{M}\right) \quad \ldots\ldots\ldots\ldots\ldots(16)$$

where C is a user-selected enhancement parameter, with effective range 0 = C < 256, M is the maximum value of the range, and μ(x, y) is the local mean of the image.

**Step 2 :**

To enhance the contrast a high-pass filter is applied on the image. This image is called $I_{EN}$. Similarly an application of edge detection algorithm is applied to the another copy of the image obtained from step1, resulting in a second image which is referred as $I_{ED}$.

**Step 3:**

Finally, the following formula gives the output-enhanced image:

$$I_{FEN} = A(I(x, y) + I_{ED}(x, y)^{\gamma} \times I_{EN}(x, y)^{\alpha}) \ldots \ldots \ldots \ldots \ldots \ldots \ldots \ldots \ldots (17)$$

Where $I_{FEN}$ is the output image and A, α, and γ are user defined operating parameters. Selection of M and C can be done simply and quickly by a human; however, selecting α and γ is a more time-consuming task.(,M = 260 and C = 50). This method preserves the local content of the input image in the output image better than HVS method .

Since the local content preservation is of utmost concern in the case of medical images, new methods for the same need to introduced .In the next section , an extension of this HVS method referred as HVSEDBI is proposed, which preserves the local content better than the former above mentioned methods. The proposed HVSEDBI method is a variation of the classical HVS method . Even though the classical HVS method is a good method for enhancement ,it does not preserve the local content of an image . Further more an another extension of this method called HVSedge ,though preserves the content better than classical HVS method ,does not preserve the contents much in the case of images like medical images where local content preservation is of utmost importance. Therefore HVSEDBI is introduced here. This method uses the enhancement scheme for dark blurred images along with the edge preserving contrast enhancement method. The proposed method is applied to different category of grey scale images like medical images and other standard images .Since HVSEDBI focuses on preserving the content of the image , a comparative study of HVSEDBI with another method referred as 'Multilayered Contrast Limited adaptive Histogram Equalization for local content preservation' is done.

## 3.PROPOSED METHODS

### 3.1. HUMAN VISION THRESHOLDING WITH ENHANCEMENT FOR DARK BLURRED IMAGES FOR LOCAL CONTENT PRESERVATION (HVSEDBI)

Another extension of the HVS method is proposed here. In this method the segmentation of the original image into 3 images is done by means of human visual system based(or human vision ) thresholding. In addition to this after the three images are obtained using the thresholding (Im1 ,Im2 and Im4) , the second image is enhanced by using normal histogram equalization , the third image using the edge preserving contrast enhancement method( mentioned in the previous section) and the third image is enhanced by using the following enhancement scheme .

In this enhancement scheme there are mainly 2 steps applied to the first image. The first step is to apply unsharp masking to the original image to get a sharper and more detailed image. Second step is contrast enhancement step that is based on a mapping function. In the following section the description of these steps are given in detail.

The steps of this enhancement technique is as follows:

**Step 1:**

**Unsharp masking step**: This step enhances the small structures and bring out the hidden details in the image by using unsharp masking. It only sharpens the areas, which have edges or large amount of details. Unsharp masking is usually performed by first generating a blurred copy of the original image by using laplacian filter and then subtracting it from the original image. This is given by the following formula .

$$I(i, j) = I_O(i, j) - I_b(i, j) \quad \ldots \ldots \ldots \ldots \ldots \ldots \ldots \ldots \ldots \ldots \ldots (18)$$

where $I(i, j)$ is the unsharp masking image, $I_O(i, j)$ is the original image, and $I_b(i, j)$ is the blurred copy.

Here before passing to the contrast enhancement step the unsharp masked image is multiplied by a value, and added to the original image to get the image that will be contrasted. Here, the large features are not altered much, but the smaller ones are enhanced. The result is a sharperand more detailed image. This operation is given by the following formula .

$$g(i, j) = I(i, j) + k\, I_O(i, j) \quad \ldots \ldots \ldots \ldots \ldots \ldots \ldots \ldots (19)$$

where $g(i, j)$ is output image, k is the scaling constant. Values for k is taken between 0.2 and 0.7

**Step 2:**

**Contrast enhancement step**: In this step, a sliding map window(3x3 window) is moved from the left side to the right side of original image horizontally starting from the image's upper right corner. A pixel value in the enhanced widow depends only on its value. This means that if the pixel of interest exceeds a certain threshold value, its value remain unchanged but if the value of the pixel is under the threshold then it will be remapped to a new value . The process can be described with the mapping function M defined below.

$$O = M(i) \quad \ldots \ldots \ldots \ldots \ldots \ldots \ldots \ldots \ldots \ldots \ldots (20)$$

where O and i are the new and old pixel values, respectively. The form of the mapping function M that determines the effect of the operation is

$$M = i*c/1 + e^{-i} \quad \ldots \ldots \ldots \ldots \ldots \ldots \ldots \ldots \ldots \ldots (21)$$

According to above mapping function the new value of corresponding pixel will be:

$$O = \begin{cases} i & \text{if } i > t \\ i + \left(i * \left(\dfrac{c}{1 + e^{-i}}\right)\right) & \text{if } i < t \end{cases} \quad \ldots \ldots \ldots \ldots \ldots \ldots (22)$$

where c is a contrast factor which determines the degree of the contrast that is needed . After map window reaches the right side of the image, it returns to the left side and moves down a step below. The process is repeated until the sliding window reaches the right-bottom corner of the entire image. On careful observation of the results, it is seen that local content preservation by this method is far better than the HVS and HVSedge method . The enhancement of contrast by this method is also comparatively better.

## 3.2 Multilayered Contrast limited adaptive histogram equalization using Frost filter (MCLAHEFROST)

This method is an extension of a local histogram equalization method referred as "Multiple layers block overlapped Histogram Equalization " (MLBOHE)[17] .As in MLBOHE there are 3 stages here ,which is the enhancement stage ,noise reduction stage and merging stage.In the enhancement stage, MLBOHE uses block overlapped histogram equalization method (BOHE) for the enhancement of images. Unlike MLBOHE ,MCLAHEFROST uses Contrast limited adaptive Histogram Equalization method (CLAHE)[17]. CLAHE is a generalization of LHE method. CLAHE method is proposed in order to overcome the problem associated with BOHE, which is the amplification of the noise level, especially in medical images . This limitation can be tackled if the enhancement rate can be controlled. In CLAHE, the enhancement rate is controlled by restricting the slope of the mapping function. A clipping limit is applied to the contextual region's histogram to reduce the undesired noise amplification and edge-shadowing effect .This is better suitable for medical images.

There is also a change in filter usage in MCLAHEFROST in comparison to MLBOHE. While MLBOHE uses median filter in the second stage , MCLAHEFROST uses Frost filter . The 3 stages are given in detail .

The aims of the Enhancement stage are to improve the contrast, to reveal the hidden details, and to redistribute the brightness evenly over the image. In order to achieve this, Stage 1 of MCLAHEFROST generates three versions of CLAHE enhanced image from an input image X. If the image has the dimensions of M XN, the dimensions of the windows used for CLAHE(i.e. M XN) are set to M/2 XN/2 M/4XN/4 ,M/8XN/8, respectively. These three CLAHE enhanced versions are denoted as h1, h2, and h3.It assumes that the brightness distribution of h1 is better than input image X. So, h1 can be used to redistribute the brightness in the image. I alsot assumes that h2 enhances the small objects, hence it can be used to reveal the hidden details. Then, it is assumed that h3 emphasizes the borders of the objects inside the image. So, h3 can be used to sharpen the image.

CLAHE tends to enhance the noise level in its output images. Thus, the aim of Noise reduction stage is to reduce the noise level from the CLAHE enhanced images. Apparently the noise generated by CLAHE is speckle noise. .Since speckle noise is a multiplicative noise , median filter will only be able to remove a less amount of such a kind of noise .So instead of using that we use a filter that is specifically used for the reduction of speckle noise .It is referred to as the frost filter

The Frost filter replaces the pixel of interest with a weighted sum of the values within the nxn moving kernel. The weighting factors decrease with distance from the pixel of interest. The weighting factors increase for the central pixels as variance within the kernel increases. This filter assumes multiplicative noise and stationary noise statistics and follows the following formula:

$$DN = \sum_{nXn} kae^{-\alpha |t|}$$

Where

$$\alpha = (4/n\sigma'^2)(\sigma^2/I'^2)$$

..............................................................(23)

K-normalization constant
I'-local mean
$\sigma$ =local variance
$\sigma'$ -image coefficient of the variation value
$|t|=|X-X_0|+|Y-Y_0|$
n-kernal size

Here we are applying the frost filter to each of the different versions of CLAHE images obtained in the previous stage so that significant amount of speckle noise is removed from each image. Therefore, in this stage, h1, h2, and h3 will be filtered with frost filters. The outputs are denoted as h1S, h2S, and h3S.

The aim of the merging stage is to produce an output image without saturation artifacts. In order to reduce this problem the mean intensity value of the enhanced image should not deviate much from the mean intensity of the input image. An approach is proposed to combine the enhanced images from Stage 2 back to the input image X. First, all the CLAHE enhanced versions are merged to become one image, P, as defined by the following equation:

P=w1*h1S+w2*h2S+w3*h3S .......................................................................(24)

where w1, w2 and w3 are weighting factors and in order to make these values image dependent these values are proportional to the entropy values as given below An entropy of the image is given by the following equation

$$E = -\sum_{x=0}^{L-1} p(x) \log_{10} p(x)$$ ....................................................................(25)

w1 w2 and w3 are chosen in such a way that it satisfies the following equation

$$w1:w2:w3 :: \left| E_{h_{1S}} - \lfloor E_x \rfloor \right| : \left| E_{h_{2S}} - \lfloor E_x \rfloor \right| : \left| E_{h_{3S}} - \lfloor E_x \rfloor \right|$$ ....................................(26)

Where $E_{h_{1S}}, E_{h_{2S}}, E_{h_{3S}}$ refers to the entropy of images ,h2S and h3S and Ex refers to the entropy of input image. This will enable the version with the highest entropy difference value to have the highest weighting value. In the merging process at the point where image P is obtained after using the corresponding weights in combination with the speckle noise reduced images as mentioned in the previous section we apply a median filter[3X3] there so that any impulse noise other than the speckle noise if present is removed so that we might get a better output.Then, in order to maintain the shape of the histogram, the final output image Y is given by:

Y =alpha*X + beta*P..................................................................(27)

where alpha and beta are positive weighting coefficients (alpha + beta=1)

MCLAHEFROST is further extended to colour images. This is done by applying the method on the red ,green and blue components of an image .Since this proposed method is only suitable for local content preservation a modification is made to it so that it can be used as a method for contrast enhancement .For this the proposed MCLAHEFROST method is combined with various concepts of existing methods as given below.

### 3.2.1 Multilayered Contrast limited adaptive Histogram Equalization with Multi histogram equalization (MCLAHEMHE)

Here the proposed MCLAHEFROST method is combined with the following method .To surmount drawbacks of normal Histogram equalization methods, the main idea used here is to decompose the image into several sub-images, such that the image contrast enhancement provided by the HE in each sub image is less intense, leading the output image to have a more natural look.Here the image decomposition process is based on the histogram of the image. The histogram is divided into classes, determined by threshold levels, where each histogram class represents a sub-image. The number of sub images into which the original image should be decomposed on depends on how the image is decomposed.In order to enhance contrast, preserve brightness and produce natural looking images,here a method referred to as Multi-HE (MHE) technique [13] which first decomposes the input image into several sub-images, and then applies the classical HE process to each of them is done here. Here a function is used to decompose the image, conceiving a method MHE method for image contrast enhancement, *i.e.*, Minimum Within-Class Variance MHE (MWCVMHE). Here the function is used to decompose an image based on threshold levels, whereas an algorithm used to find the optimal threshold levels is specified along with it . A criterion for automatically selecting the number of decomposed sub-images is also taken .A cost function, taking into account both the discrepancy between the input and enhanced images and the number of decomposed sub-images, is used to automatically make the decision of in how many sub-images the input image will be decomposed on.

**(a)Multi-Histogram Decomposition**

Here, decomposing an image is done in such a way that the enhanced images still have a natural appearance. For that clustering the histogram of the image into classes is done , where each class corresponds to a sub-image. By doing that, minimization of the brightness shift yielded by the HE process into each sub-image occurs. With the minimization of this shift, this method is expected to preserve both the brightness and the natural appearance of the processed image. From the multi-threshold selection literature point of view,the problem stated above can be seen as the minimization of the within-histogram class variance , where the within-class variance is the total squared error of each histogram class with respect to its mean value (*i.e.*, the brightness). That is, the decomposition aim is to find the optimal threshold set $T_k = \{t_1^k\ t_2^k\ t_3^k\ t_{k-1}^k\}$ which minimizes the decomposition error of the histogram of the image into $k$ histogram classes and decomposes the image $I[0, L-1]$ into $k$ sub-images $I[ls^{1,k}, lf^{1,k}]$….. $I[ls^{k,k}, lf^{k,k}]$ where $ls^{j,k}$ and $lf^{j,k}$ are lower and upper gray-level boundaries of each sub-image $j$ when the image is decomposed into $k$ sub-images. They are defined as $ls^{j,k} = t_{j-1}^k$ if $j>1$ and $ls^{j,k}=0$ otherwise and $lf^{j,k} = t_j^k+1$ if $j$ not equal to $k$ and $lf^{j,k}$ is zero otherwise .The discrepancy function for decomposing the original image into $k$ sub-images following the minimization of within-class variance can be expressed as

$$Disc(k) = \sum_{j=1}^{k} \sum_{l=l_s^{j,k}}^{l_f^{j,k}} \left(l - l_m\left(I\left[l_s^{j,k}, l_f^{j,k}\right]\right)\right)^2 P_l^{[0,L-1]} \quad \ldots\ldots\ldots\ldots\ldots \quad (28)$$

The method conceived with this discrepancy function will be called Minimum Within-Class Variance MHE (MWCVMHE). Note that the mean gray-level (*i.e.*, the brightness) of each sub-image processed by the CHE method is theoretically shifted to the middle gray-level of its range, *i.e.*, $lm(O[ls, lf]) = lmm(I[ls, lf]) = (ls + lf)/2$.

**(b) Algorithm for computing the threshold matrix**

1. *for* $q \leftarrow 0; q < L; q++ do D(l)_q \leftarrow \phi(0, q)$;
2. for $p \leftarrow L; p \leq k; p++$ do
3. $D(p+1)_p \leftarrow D(p)_{p-1} + \phi(p-1, p-1)$;
4. $PT(p+1, p) \leftarrow p-1$;
5. *for* $q \leftarrow p+1; q \leq L-k+p; q++ do$
6. $D(p+1)_q \leftarrow -\infty$;
7. for $l \leftarrow p-1; l \leq q-1; l++$ do
8. if $(D(p+1)_q > D(p)_l + \phi(l+1, q))$ *then*
9. $D(p+1, q) \leftarrow D(p)_l + \phi(l+1, q)$;
10. $PT(p+1, q) \leftarrow l$;

Here ɸ ( $p,q$) - discrepancy of sub-image $I(p, q)$ , $D(p)q$ - disc. function $Disc(p)$ up to level $q$, $PT$ - optimum thresholds matrix

**(c) Finding the Optimal Thresholds**

The task of finding the optimal $k-1$ threshold levels which segment an image into $k$ classes can be easily obtained from the threshold matrix by using the following conditions
The threshold from threshold matrix PT is obtained as

$$t_k^{j=} = PT(j+1, t_{j+1}^{k*}) \quad \ldots\ldots\ldots\ldots\ldots\ldots(29)$$

where $1<=j<k$, $t_{j+1}^{k*}=L-1$ if $j+1=k$ and $t_{j+1}^{k*} = t_{j+1}^{k}$ otherwise

**(d) Automatic Thresholding Criterion**

This section presents an approach to automatically choose in how many sub-image the original image should be decomposed on. Fro this here a cost function is used . This cost function takes into account both the discrepancy between the original and processed images and the number of sub images to which the original image is decomposed, and it is defined as

$$C(k) = \rho\,(Disc(k)) + (\log 2\, k) \quad \ldots\ldots\ldots\ldots\ldots\ldots(30)$$

where ρ is a positive weighting constant. The number of decomposed sub-images $k$ is automatically given as the one which minimizes the cost function $C(k)$.

### 3.2.2 Multilayered Contrast limited adaptive Histogram Equalization with Enhancement for dark Blurred images (MCLAHEEDBI)

Here we combined the proposed MCLAHEFROST method with the enhancement scheme for blurred images mentioned in section 3.1.

### 3.2.3 .Multilayered Contrast limited adaptive Histogram Equalization using frost filter with adaptive local region stretching (MCLAHEALRS)

Here the division of the histogram of the luminance levels of the output image of MCLAHEFROST into three regions – dark, mid and bright is done . These regions are of equal size. Each of these three regions is then processed independently using HE. The effect of HE is then toned down depending on the shape of the histogram of each region. This independent processing has two main advantages. Once HE has been performed independently for the three regions, the final output is obtained by taking a weighted average of the input with the HE output. This weighting factor is independently calculated and controlled for the three regions. There are two reasons for performing this weighted average. Firstly, it can be used to control the level of enhancement and is well suited to program modes of operation such as low, medium and high settings. Secondly, and more importantly, it can be used to adjust the level of enhancement differently for the three regions. The weighting factor for each region of the histogram is calculated from the pseudo-variance of each region. First the mean luminance value of each region is found. Then the mean value is used to calculate the pseudo-variance using the following formula .

$$\sigma i = \frac{1}{N}\sum_{j} nj(|\,yj - mi\,|) \quad \ldots\ldots\ldots\ldots\ldots\ldots(31)$$

where '$mi$' is the mean luminance value, '$nj$' is the number of points at luminance level '$yj$', N is the total number of points in the region and $\sigma i$ is the pseudo-variance of the region. The above summation is carried out over all pixels belonging to one region and subscript i Є [1, 3] refers to the region being processed.

## 4. RESULTS AND DISCUSSIONS

For the experiment several categories of medical images are being used .Here an analysis based on brain images , mammograms and a knee images are given .In addition to this the result of application of this method to the well known images like 600 × 600 pixel Einstein ,girl image and other images are also specified .For EPCE used for enhancing one of the thresholded images, A is taken as 40,gamma=1,alpha=1,C=240,M=255.Generally for HVS thresholding $\alpha_2$=0.1, $\alpha_1$=0, $\alpha_3$=0,beta= -1.5. The alpha and beta values for MCLAHEFROST method are taken as 0.9 and 0.1 and w1 ,w2 ,w3 takes up the values 0.7,0.2, 0.1 for obtaining the following results.The performance measures used here are Peak Signal to Noise Ratio (PSNR)and Average mean brightness error (AMBE) and Contrast improvement index(CII).The measure AMBE is mainly taken because it is used to specify the local content preservation .So if AMBE value obtained by a particular method is lesser compared to other methods , then the former method preserves the local content of the input image in the output image better. Generally the PSNR value of a image R with respect to the original image O, both of size M × N, is calculated as shown below.

$$PSNR = 10 * \log 10 (255^2 / MSE) \quad\quad\quad\quad (32)$$

Where MSE is the mean squared error given by

$$MSE = \sum_{m,n} [O(m,n) - R(m,n)]^2 / (M*N)$$
$$\quad\quad\quad\quad (33)$$

The Average mean brightness error (AMBE) is basically used to measure the capability of the method to maintain the mean brightness of the input image in its output image.

The AMBE is given by: AMBE =|X'-Y'| ….............................(34)

where X' and Y' are the average intensity values of the input and output image, respectively. The brightness of the input image should be retained in the output image in order to avoid intensity saturation problem. Hence, a good enhancement method should have a low value of AMBE.

Contrast improvement index (CII) is used to measure the increase in the contrast generated by the methods .For this a 3X3 window is taken and passed over the entire image . At each window the maximum and minimum intensity value is found . The summation of all the maximum values and the summation of all the minimum values is obtained for the entire image. Then the ratio of (maximum-minimum) and (maximum+minimum) is taken and multiplied by 1/ number of windows. Finally the ratio of this above obtained value for the input and output image gives the CII value for the image obtained by the corresponding method .

**Table 1** Comparison of the AMBE values produced by (HVS) and HVSedge and the proposed method HVSEDBI

| *Image* | *AMBE* | | |
|---|---|---|---|
| | **HVS** | **HVSedge** | **HVSEDBI** |
| **Lena** | 34.75 | 26.78 | **22.78** |
| **Einstein** | 27.91 | 19.92 | **15.92** |
| **Couple** | 21.17 | 13.23 | **9.22** |
| **Girl** | 17.04 | 9.04 | **5.04** |
| **House** | 17.88 | 9.97 | **5.91** |
| **Clock** | 10.03 | 2.10 | **1.88** |
| **Peppers** | 25.46 | 14.21 | **9.6** |

From Table1 it is observed that the proposed method (HVSEDBI)preserves the local content of the input image in the output image better than the existing methods. In all the images a significant decrease in the AMBE values is observed making the proposed method better than the existing methods for local content preservation .

**Table 2** Comparison of the PSNR values produced by (HVS) and HVSedge and the proposed method HVSEDBI

| Image | PSNR | | |
|---|---|---|---|
| | HVS | HVSedge | HVSEDBI |
| **Lena** | 25.28 | 25.95 | **26.30** |
| **Einstein** | 25.64 | 26.31 | **26.70** |
| **Couple** | 26.40 | 26.98 | **27.30** |
| **Girl** | 26.09 | 28.31 | **29.52** |
| **House** | 26.45 | 27.30 | **27.88** |
| **Clock** | 27.90 | 30.54 | **31.62** |
| **Peppers** | 25.94 | 26.67 | **27.26** |

From the above table (table2) it is observed that the the PSNR values obtained for the proposed method is better compared to the values obtained for the existing methods indicating that the proposed method generates images of better quality compared to the existing methods. The following images (Figure 1) shows examples of outputs produced by the existing methods and the proposed method((HVSEDBI) for standard images .

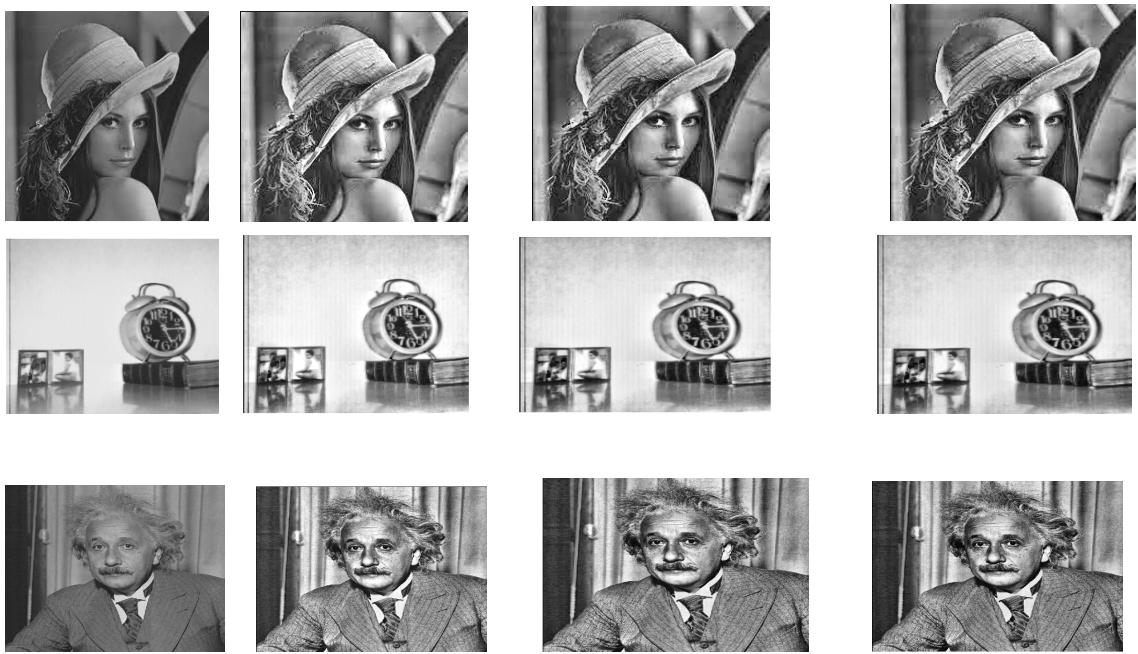

(a) (b) (c) (d)

**Figure.1** Column(a) Original Images Lena ,Clock ,Einstein Column (b) HVS Column (c) Output using HVSedge Column (d) Output using HVSEDBI

Since local content preservation is a major factor in the case of medical images ,the proposed method (HVSEDBI) is also applied on medical images .The table below gives the results obtained likewise .

**Table 3** Comparison of the AMBE values produced by (HVS) and HVSedge and the proposed method HVSEDBI for medical images

| Image | AMBE | | |
|---|---|---|---|
| | HVS | HVSedge | HVSEDBI |
| Knee1 | 21.43 | 13.47 | **7.27** |
| Knee4 | 27.51 | 19.53 | **14.15** |
| Mammo1 | 26.00 | 18.02 | **12.77** |
| Mammo2 | 35.36 | 27.37 | **12.68** |
| Brain1 | 18.86 | 12.86 | **6.18** |
| Brain2 | 24.49 | 18.51 | **11.70** |
| Brain3 | 23.05 | 17.05 | **9.5** |
| Brain4 | 21.62 | 15.63 | **8.08** |
| Mammo0 | 27.66 | 19.67 | **15.34** |

As mentioned before ,a method is said to preserve the local content of an input image in the output image better if the AMBE value generated by it for an image is low .From the above table it is seen that there is significant decrease in the AMBE value generated by the proposed method compared to the existing methods which in turn is very essential in the case of medical images .

**Table 4** Comparison of the PSNR values produced by (HVS) and HVSedge and the proposed method HVSEDBI for medical images

| Image | PSNR | | |
|---|---|---|---|
| | HVS | HVSedge | HVSEDBI |
| Knee1 | 25.21 | 26.21 | **28.28** |
| Knee4 | 24.84 | 26.01 | **27.99** |
| Mammo1 | 25.25 | 26.33 | **27.78** |
| Mammo2 | 24.31 | 24.79 | **28.97** |
| Brain1 | 25.76 | 27.26 | **30.18** |
| Brain2 | 25.43 | 26.74 | **29.33** |
| Brain3 | 25.41 | 26.76 | **29.64** |
| Brain4 | 25.42 | 26.76 | **29.96** |
| Mammo0 | 24.91 | 26.94 | **28.02** |

From the above table it can be observed that is a significant increase in the PSNR values obtained by the proposed methods (HVSEDBI) compared to the existing methods for medical images also.The following images (Figure 2) shows examples of outputs produced by the existing methods and the proposed method ((HVSEDBI) in the different category of medical images .

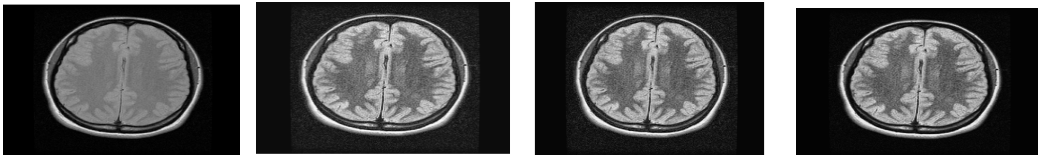

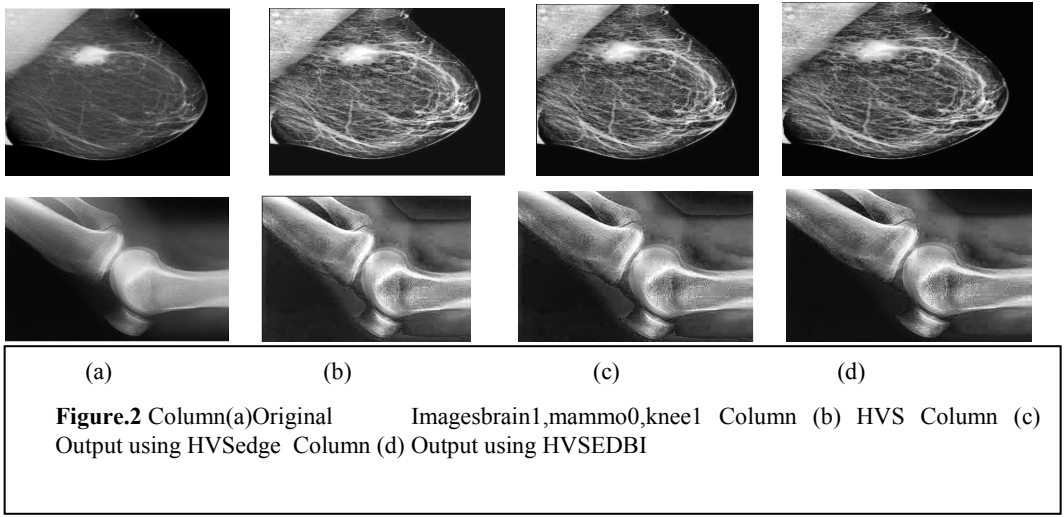

**Figure.2** Column(a)Original Imagesbrain1,mammo0,knee1 Column (b) HVS Column (c) Output using HVSedge Column (d) Output using HVSEDBI

Finally since the proposed methods and the existing methods used for comparison are enhancement methods, a comparison of the enhancement in the contrast obtained by the above methods using a performance measure called contrast improvement index(CII)is shown in Table 5.

**Table 5** Comparison of the CII values produced by (HVS) and HVSedge and the proposed method HVSEDBI for standard images

| Image | CII | | |
|---|---|---|---|
| | **HVS** | **HVSedge** | **HVSEDBI** |
| **Lena** | 1.51 | 1.6 | **1.74** |
| **Einstein** | 2.2 | 2.4 | **2.53** |
| **Couple** | 2.0 | 2.1 | **2.2** |
| **Girl** | 2.2 | 2.4 | **2.5** |
| **House** | 1.52 | 1.6 | **1.79** |
| **Clock** | 1.63 | 1.76 | **1.84** |
| **Peppers** | 1.06 | 1.51 | **1.56** |

**Table 6** Comparison of the AMBE values produced by (HVS) and HVSedge and the proposed methods HVSEDBI and MCLAHEFROST for standard images

| Image | AMBE | | | |
|---|---|---|---|---|
| | **HVS** | **HVSedge** | **HVSEDBI** | **MCLAHEFROST** |
| **Lena** | 34.75 | 26.78 | 22.78 | **5.19** |
| **Einstein** | 27.91 | 19.92 | 15.92 | **4.47** |
| **Couple** | 21.17 | 13.23 | 9.22 | **3.1** |
| **Girl** | 17.04 | 9.04 | 5.04 | **2.3** |
| **House** | 17.88 | 9.97 | 5.91 | **3.3** |
| **Clock** | 10.03 | 2.10 | 1.88 | **0.35** |
| **Peppers** | 25.46 | 14.21 | 9.6 | **3.73** |

**Table 7** Comparison of the PSNR values produced by (HVS) and HVSedge and the proposed methods HVSEDBI and MCLAHEFROST for standard images

| Image | PSNR | | | |
|---|---|---|---|---|
| | HVS | HVSedge | HVSEDBI | MCLAHEFROST |
| **Lena** | 25.28 | 25.95 | 26.30 | **32.46** |
| **Einstein** | 25.64 | 26.31 | 26.70 | **33.31** |
| **Couple** | 26.40 | 26.98 | 27.30 | **35.12** |
| **Girl** | 26.09 | 28.31 | 29.52 | **39.75** |
| **House** | 26.45 | 27.30 | 27.88 | **33.14** |
| **Clock** | 27.90 | 30.54 | 31.62 | **38.60** |
| **Peppers** | 25.94 | 26.67 | 27.26 | **33.11** |

From the above tables it can be observed that the proposed methods preserve the contents better compared to the existing methods for standard images .In addition to that the proposed methods (HVSEDBI and MCLAHEFROST) gives better values compared to the existing methods.In comparison of the proposed methods ,MCLAHE FROST outperforms HVSEDBI in AMBE and PSNR value**s** for standard images **.**The following images shows examples of outputs produced by the existing methods and the two proposed methods ((HVSEDBI and MCLAHEFROST) in the following standard images .

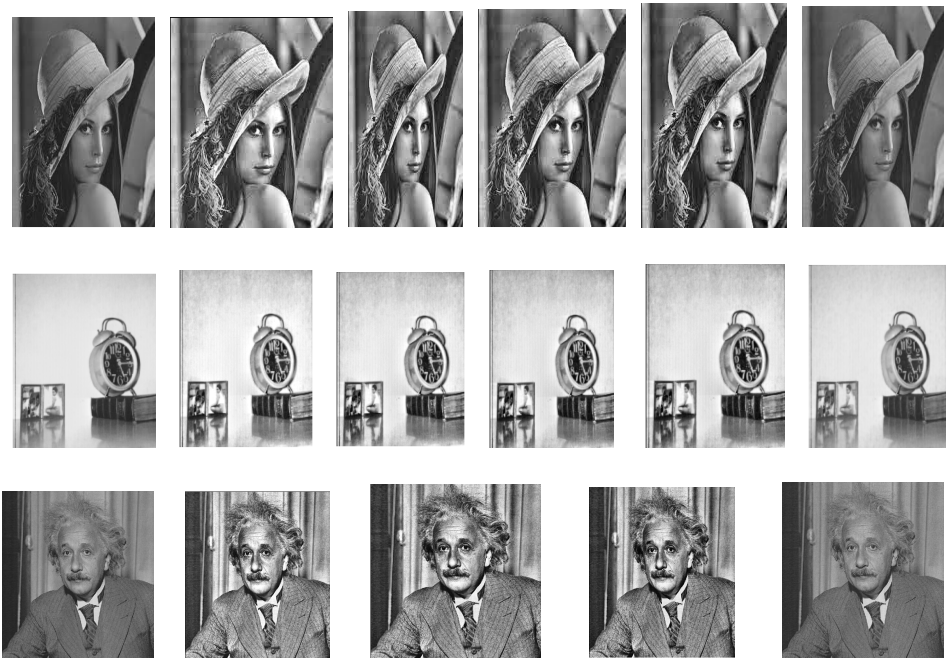

(a)     (b)     (c)     (d)     (e)

**Figure.3** Column(a)Original Images Lena ,Clock ,Einstein Column (b) HVS Column (c) Output using HVSedge Column (d) Output using HVSEDBI (e) ) Output using MCLAHEFROST

**Table 8** Comparison of the AMBE values produced by (HVS) and HVSedge and the proposed methods HVSEDBI and MCLAHEFROST for medical images

| Image | AMBE | | | |
|---|---|---|---|---|
| | HVS | HVSedge | HVSEDBI | MCLAHEFROST |
| **Knee1** | 21.43 | 13.47 | 7.27 | **5.5** |
| **Knee4** | 27.51 | 19.53 | 14.15 | **5.7** |
| **Mammo1** | 26.00 | 18.02 | 12.77 | **5.5** |
| **Mammo2** | 35.36 | 27.37 | 12.68 | **4.8** |
| **Brain1** | 18.86 | 12.86 | 6.18 | **3.8** |
| **Brain2** | 24.49 | 18.51 | 11.70 | **7.2** |
| **Brain3** | 23.05 | 17.05 | 9.5 | **7.4** |
| **Brain4** | 21.62 | 15.63 | 8.08 | **4.0** |
| **Mammo0** | 27.66 | 19.67 | 15.34 | **5.5** |

From table 8 it can be observed that the proposed methods preserve the contents better compared to the existing methods for medical images and MCLAHE FROST outperforms HVSEDBI in AMBE value**s** for medical images too **.**The following images shows examples of outputs produced by the existing methods and the two proposed methods ((HVSEDBI and MCLAHEFROST) in the following medical images .

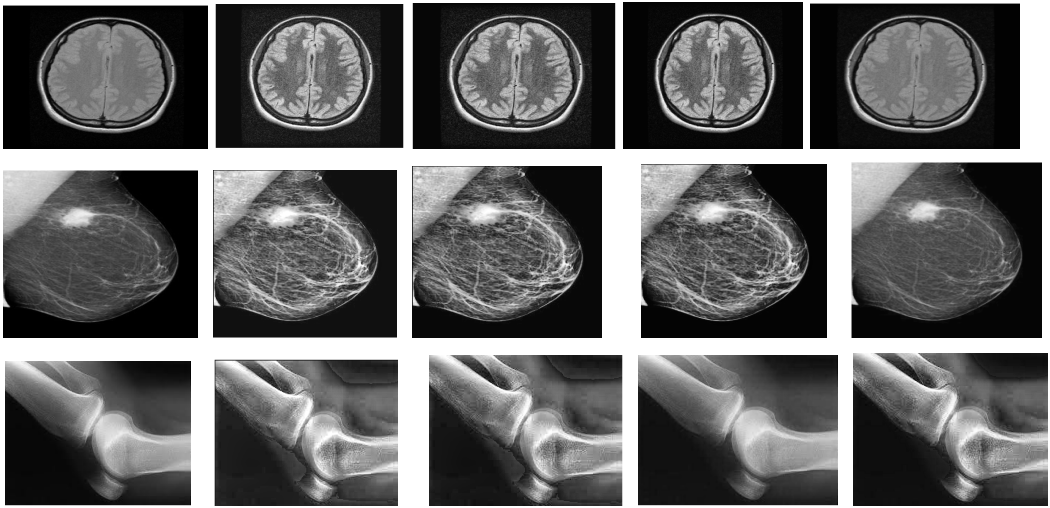

(a)      (b)      (c)      (d)

**Figure .4** Column(a)Original     Image s brain1,mammo0,knee1 Column (b) HVS Column (c) Output using HVSedge   Column (d) Output using HVSEDBI (e) ) Output using MCLAHEFROST

**Table 9** Comparison of the AMBE values produced by (HVS) and HVSedge and the proposed methods HVSEDBI and MCLAHEFROST for aerial images

| Image | AMBE | | | |
|---|---|---|---|---|
| | HVS | HVSedge | HVSEDBI | MCLAHEFROST |
| 2.1.01 | 23.7 | 15.85 | 11.87 | 3.2 |
| 2.1.02 | 7.68 | 15.46 | 19.44 | 0.05 |
| 2.1.03 | 15.03 | 7.03 | 3.03 | 2.39 |
| 2.1.04 | 13.45 | 5.52 | 1.5 | 1.5 |
| 2.1.05 | 14.49 | 6.62 | 2.6 | 2.6 |
| 2.1.06 | 9.64 | 17.49 | 21.48 | 0.02 |
| 2.1.07 | 4.14 | 12.09 | 16.09 | 0.39 |
| 2.1.08 | 19.03 | 27.02 | 31.02 | 2.3 |
| 2.1.09 | 6.08 | 14.06 | 18.06 | 0.53 |

From table 9 it can be observed that the proposed methods preserve the contents better compared to the existing methods for aerial images . On comparing the proposed methods, MCLAHE FROST is better than HVSEDBI for local content preservation in aerial images also. The following images shows examples of outputs produced by the existing methods and the two proposed methods ((HVSEDBI and MCLAHEFROST) in aerial images .

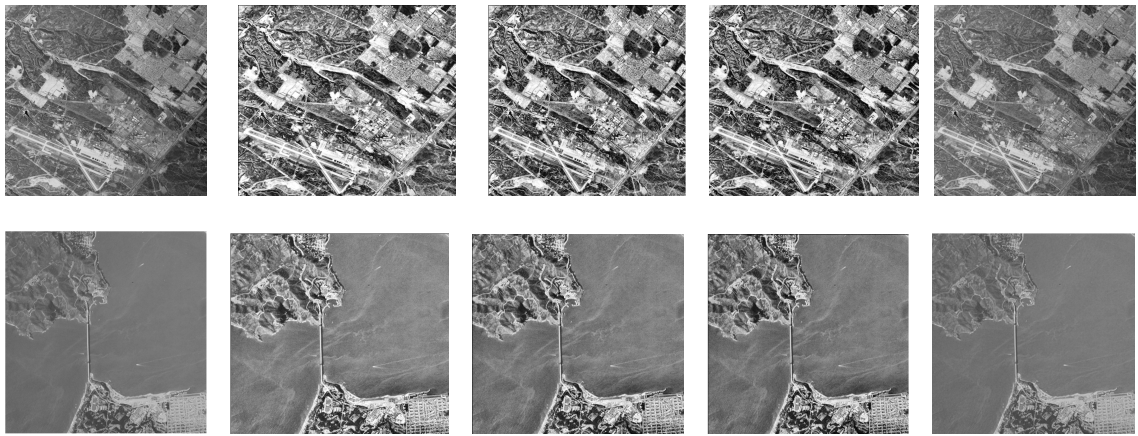

(a) (b) (c) (d) (e)

**Figure .5** Column(a)Original Images 2.1.01 and 2.1.03 Column (b) HVS Column (c) Output using HVSedge Column (d) Output using HVSEDBI (e) ) Output using MCLAHEFROST

The following shows the result of application of MCLAHEFROST method on a real image .

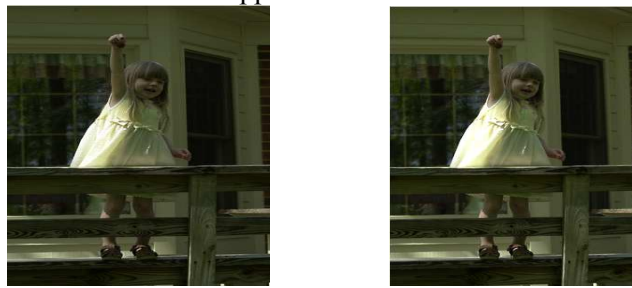

(a) (b)

**Figure. 6** Column (a) Original Image Image2 Column (b) Output using MCLAHEFROST

From the above image it can be observed that with MCLAHEFROST even though a small amount of enhancement is seen visually ,increase in contrast does not occur here. So inorder to make it a good contrast enhancement method , an incorporation of various concepts of several existing methods with the proposed MCLAHEFROST method is done . The results of the same are given below .

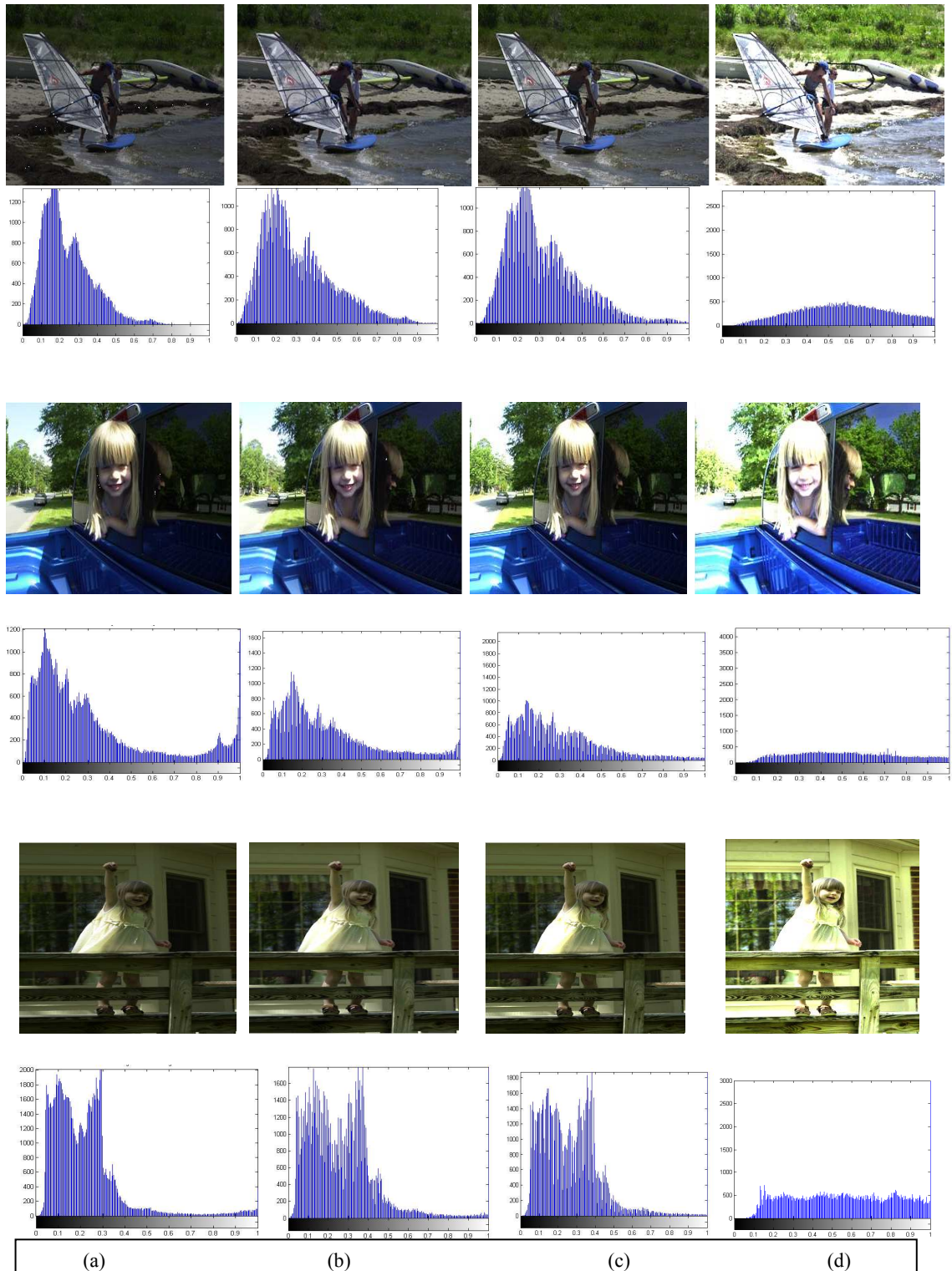

(a)      (b)      (c)      (d)

**Figure. 7**. Column (a) Original Image Image7,Image7histogram  Image8, Image8 histogram ,Image2, Image2 histogram Column (b) Output using MCLAHEMHEColumn (c) Output using MCLAHEEDBI (d)Output using MCLAHEALRS

From the figures and histograms it can be observed that among the three proposed methods MCLAHEALRS outperforms others in contrast improvement .

**Table 10** Comparing the CII values of proposed methods MCLAHEMHE , MCLAHEEDBI,MCLAHEALRS

| CII | MCLAHEMHE | MCLAHEEDBI | MCLAHEALRS |
|---|---|---|---|
| Image2 | X | **1.27** | **2.06** |
| Image3 | X | **1.21** | **1.59** |
| Image4 | X | **1.16** | **1.79** |
| Image6 | X | **1.27** | **1.3** |
| Image7 | 1.1 | **1.38** | **2.91** |
| Image8 | X | **1.26** | **1.12** |
| Image9 | 1.12 | **1.31** | **2.44** |
| Image11 | 1.81 | **1.87** | **5.48** |
| pic1 | X | **1.12** | **1.18** |
| pic4 | X | **1.16** | **1.8** |

From the above table it can be seen that when the concepts of multi histogram equalization was incorporated in MCLAHEFROST (ie in method MCLAHEMHE)only in the images Image 7,Image 9 ,Image11 the increase in contrast is observed .The rest of the images (ie, the ones marked 'X') no improvement is seen .In comparison to MCLAHEMHE , MCLAHEEDBI gives a good increase in contrast for all the images .Finally in MCLAHEALRS a significantly better increase in contrast is seen for all the images

## 5.CONCLUSION

In this paper , a HVS based image enhancement method is introduced. A number of enhancement algorithms can be used with HVS based methods .Here an enhancement algorithm for dark blurred images is used with H uman Vision thresholding (HVSEDBI) mainly for the preservation of local content of the images . Furthermore a comparitive study with another method referred as MCLAHEFROST, for local content preservation is also done .Experiments on real medical images such as brain, mammogram images and knee images shows the superiority of the proposed methods (HVSEDBI and MCLAHEFROST )over the existing methods (HVS and HVSedge ) in local content preservation.The results obtained for standard images other than the medical images shows that the proposed method is even efficient for those images also .While HVSEDBI is a good method for local content preservation and contrast enhancement , MCLAHEFROST is a better method for giving a natural look to the image and comparatively better AMBE values for standard ,medical and aerial images. Furthermore MCLAHEFROST is extended to colour images. Even though MCLAHEFROST is a good method for local content preservation ,it does not increase the contrast of the image much .So it is combined with the concepts of several existing methods to generate the methods MCLAHEMHE,MCLAHEEDBI and MCLAHEALRS. From the experiment results it is observed that MCLAHEALRS gives good results qualitatively and quantitatively .

## REFERENCES

[1]. A. Rosebfield, A. C. Kak, Digital Picture processing,Academic press, San Diego, CA, 1976.

[2]. R. C. Gonzalez, Richard E. Woods, Digital Imageprocessing, Addision-Wesely, 2003

[3] R. H. Sherrier, G. A. Johnson, "Regionally adaptive histogram equalization of the chest," *IEEE Trans.Med.Image*, Mi-6(1987), pp.1-7.

[4] S. M. Pizer, E. P. Amburn, "Adaptive histogram equalization and its variations," *Compt. Vision, Graph, Image Process*,39(1987), pp. 355-368.

[5] Md. Foisal Hossain, Mohammad Reza Alsharif"Image Enhancement Based on Logarithmic Transform Coefficient and Adaptive Histogram Equalization", 2007 International Conference on Convergence Information Technology.

[6] Petrou Maria, Bosdogianni Panagiota. Image Processing: The Fundamentals. John Wiley and Sons; 1999.

[7] Zimmerman John B, Pizer Stephen M, Staab Edward V, Randolph Perry J, McCartney William, Brenton Bradley C. An evaluation of the effectiveness of adaptive histogram equalization for contrast enhancement. *IEEE Trans Med Imag* 1988;7(4):04–312.

[8] Kim Joung-Youn, Kim Lee-Sup, Hwang Seung-Ho. An advanced contrast enhancement using partially overlapped sub-block histogram equalization.*IEEE Trans Circuits Syst Video* Technol 2001;11(4):475–84.

[9] Pizer Stephen M, Philip Amburn E, Austin John D, Cromartie Robert, Geselowitz Ari, Greer Trey, et al. Adaptive histogram equalization and itsvariations. Comput Vision Graph Imag Process 1987;39(3):355–68.

[10] Pizer Stephen M, Johnston R. Eugene, Ericksen P. James, Bonnie C. Yankaskas, Muller Keith E. Contrast-limited adaptive histogram equalization speed and effectiveness. In: Proceedings of the .rst conference on visualization in biomedical computing, 1990. p. 337–45.

[11] Naglaa Yehya Hassan1, and Norio Aakamatsu"Contrast Enhancement Technique of Dark Blurred Image", *IJCSNS International Journal of Computer Science and Network Security,* VOL.6 No.2A, February 2006

[12] Karen A. Panetta, *Fellow, IEEE*, Eric J. Wharton,"Human visual system based multi histogram equalization for non uniform illumination and shadow correction",ICASSP2007

[13] David Menotti, Laurent Najman, Jacques Facon, and Arnaldo de A. Araujo , "Multi-histogram equalization method for contrast enhancement and brightness preserving", *IEEE Transactions on Consumer Electronics*, Vol. 53, No. 3, AUGUST 2007

[14] Karen A. Panetta, *Fellow, IEEE*, Eric J. Wharton, *Student Member, IEEE*, and Sos S. Agaian, *Senior Member, IEEE,*"Human visual system- based image enhancement and logarithmic contrast measure", IEEE TRANSACTIONS ON SYSTEMS, MAN, AND CYBERNETICS—PART B: CYBERNETICS, VOL. 38, NO. 1, FEBRUARY 2008

[15] David Menotti, Laurent Najman, Jacques Facon, and Arnaldo de A. Araujo , "Contrast Enhancement in Digital imaging using Histogram Equalization", Universidade Federalde Minas Gerais, in April 2008

[16] S. Srinivasan, N. Balram*,*"Adaptive Contrast Enhancement Using Local Region Stretching**, *Proc.of ASID'06, 8-12 Oct, New Delhi*

[17] Nicholas Sia Pik Kong, Haidi Ibrahim"Multiple layers block overlapped histogram equalization for local content emphasis" Computers and Electrical Engineering xxx (2010) xxx–xxx